\documentclass[journal]{IEEEtran}

\ifCLASSINFOpdf
\else
   \usepackage[dvips]{graphicx}
\fi
\usepackage{url}
\usepackage{xcolor}
\newcommand{\tabincell}[2]{\begin{tabular}{@{}#1@{}}#2\end{tabular}}

\usepackage{graphicx}
\usepackage{multirow}
\usepackage{balance}
\usepackage{float}
\usepackage{algorithmicx,algorithm}

\begin{document}

\title{
Similarity Transfer for Knowledge Distillation
}

\author{Haoran Zhao, Kun Gong, Xin Sun, \IEEEmembership{Member, IEEE}, Junyu Dong, \IEEEmembership{Member, IEEE} \\ and Hui Yu, \IEEEmembership{Senior Member, IEEE}

\thanks{This work was supported in part by the National Natural Science Foundation of China (No. U1706218, 61971388), and Major Program of Natural Science Foundation of Shandong Province (No. ZR2018ZB0852). (Corresponding author: Xin Sun and Junyu Dong.)}
\thanks{H. Zhao, K. Gong, X. Sun, and J. Dong are with the College of Information Science and Engineering, Ocean University of China, 238 Songling Road, Qingdao, China (e-mail:zhaohaoran@stu.ouc.edu.cn; sunxin@ouc.edu.cn; dongjunyun@ouc.edu.cn).}
\thanks{H. Yu is with the School of Creative Technologies, University of Portsmouth, Portsmouth, PO1 2DJ, UK (e-mail: hui.yu@port.ac.uk).}

}

\markboth{Journal of \LaTeX\ Class Files, Vol. 14, No. 8, August 2015}
{Shell \MakeLowercase{\textit{et al.}}: Bare Demo of IEEEtran.cls for IEEE Journals}
\maketitle

\begin{abstract}
Knowledge distillation is a popular paradigm for learning portable neural networks by transferring the knowledge from a large model into a smaller one. Most existing approaches enhance the student model by utilizing the similarity information between the categories of instance level provided by the teacher model. However, these works ignore the similarity correlation between different instances that plays an important role in confidence prediction. To tackle this issue, we propose a novel method in this paper, called similarity transfer for knowledge distillation (STKD), which aims to fully utilize the similarities between categories of multiple samples. Furthermore, we propose to better capture the similarity correlation between different instances by the mixup technique, which creates virtual samples by a weighted linear interpolation. Note that, our distillation loss can fully utilize the incorrect classes similarities by the mixed labels. The proposed approach promotes the performance of student model as the virtual sample created by multiple images produces a similar probability distribution in the teacher and student networks. Experiments and ablation studies on several public classification datasets including CIFAR-10,CIFAR-100,CINIC-10 and Tiny-ImageNet verify that this light-weight method can effectively boost the performance of the compact student model. It shows that STKD substantially has outperformed the vanilla knowledge distillation and has achieved superior accuracy over the state-of-the-art knowledge distillation methods.

\end{abstract}


\IEEEpeerreviewmaketitle

\section{Introduction}

\IEEEPARstart{D}{eep} convolutional neural networks (CNNs) have made unprecedented advances in a wide range of computer vision applications such as image classification \cite{Russakovsky2015ImageNet}\cite{He_2016_CVPR}, object detection \cite{Li_2017_CVPR}\cite{9201410} and semantic segmentation \cite{xie2018improving}. However, these top-performing neural networks are usually developed with large depth, parameters and high complexity, which consume expensive computing resources and make it hard to be deployed on low-capacity edge devices. With the increasing demands for low cost networks and real-time response on resource-constrained devices, there is an urgently need for novel solutions that can reduce model complexities while keeping decent performance.

To tackle this problem, a large body of works has been proposed to accelerate or compress these deep neural networks in recent years. Generally, these solutions fall into the following perspectives: network pruning \cite{Cun1989Optimal}\cite{Hao2016Pruning}\cite{DBLP:journals/corr/abs-1901-07827}, network decomposition \cite{Denil2013Predicting}\cite{Kim2015Compression}\cite{DBLP:conf/nips/RenZ0PCZL018}, network quantization and knowledge distillation \cite{Hinton2015Distilling}\cite{9269400}. Among these methods, the seminal work of knowledge distillation has attracted a lot of attention due to its ability of exploiting dark knowledge from the pre-trained large network. 

Knowledge distillation (KD) is proposed by Hinton et al.  \cite{Hinton2015Distilling} for supervising the training of a compact yet efficient student model by capturing and transferring the knowledge of a large teacher model to a compact one. Its success is attributed to the knowledge contained in class distributions provided by the teacher via soften softmax. Many follow up works have been proposed since then focusing on different categories of knowledge with various kinds of distillation loss functions in the field of CNNs optimization. However, it is still an open question to define the distillation loss, i.e. how to best capture and define the knowledge of teacher to train the student.

In vanilla KD \cite{Hinton2015Distilling}, the knowledge of teacher network is transferred to the student network by mimicking the class probabilities outputs, which are softened by setting a temperature hyperparameter in softmax. Inspired by KD, following works introduce the output of intermediate layers as knowledge to supervise the training of the student network. For example, FitNets \cite{Romero2015FitNets} first introduces the intermediate representation as the hints to guide the training of the student network, which directly matches the feature maps of intermediate layers between the teacher network and the student network. Later, Zagoruyko et al. \cite{Zagoruyko2016Paying} introduce attention maps (AT) from the feature maps of intermediate layers as knowledge. FSP \cite{Yim2017A} designs a flow distillation loss to encourage the student to mimic the teacher's flow matrices within the feature maps between two layers. Recently, Park et al. \cite{park2019relational} propose a relational knowledge distillation (RKD) method which draws mutual relations of data examples by the proposed distance-wise and angle-wise distillation losses.
In particular, SP \cite{tung2019similarity} preserves the pairwise similarities in student's representation space instead to mimic the representation space of the teacher. Besides, various approaches extend these works by matching other statistics, such as gradient \cite{pmlr-v80-srinivas18a} and distribution \cite{DBLP:conf/eccv/PassalisT18}. 

Recently, there have been some attempts to extend KD to other domains where KD also shows its potential. For example, Papernot et al. \cite{DBLP:conf/sp/PapernotM0JS16} introduce the defensive distillation for adversarial attack to reduce the effectiveness of adversarial samples on CNNs. Gupta et al. \cite{DBLP:conf/cvpr/GuptaHM16} transfer the knowledge among the images from different modalities. Besides, knowledge distillation can be employed to some task-specific methods such as 
object detection \cite{uijlings2018revisiting}, semantic segmentation \cite{DBLP:conf/cvpr/LiuCLQLW19}\cite{DBLP:conf/cvpr/JiaoWJSLH19}, face model compression \cite{luo2016face} and depth estimation \cite{DBLP:conf/cvpr/PilzerLS019}\cite{9004476}.

From a theoretical perspective, Yuan et al. \cite{DBLP:conf/cvpr/YuanTLWF20} interpret knowledge distillation in terms of Label Smoothing Regularization \cite{DBLP:conf/cvpr/SzegedyVISW16} and find the importance of soft targets regularization. They propose to manually design the regularization distribution to instead the teacher network. Notably, Hinton et al. \cite{NEURIPS2019_f1748d6b} investigate the effect of label smoothing to knowledge distillation and observe that label smoothing can alter the performance of distillation when the teacher model is trained with label smoothing. Through visualization of penultimate layer's activations of CNNs \textcolor{red}{.}, they find that the label smoothing could discard the similarity information of categories, causing poor distillation results. Thus, the similarity information of categories is vital for knowledge distillation. But this issue has been ignored by existing methods. As mentioned above, most existing works investigate knowledge distillation depending on the similarities of instance level.

Different from previous methods, we aim to exploit the privileged knowledge on similarity correlation between different instances using virtual samples, created by mix-samples. Generally speaking, the classification confidence represents over-confident predictions when the high-capacity teacher model has the excellent performance. Consequently, the incorrect classes contain low confidence which has low similarity information among categories. As shown in Fig. \ref{fig:framwork}, the teacher from the Vanilla Knowledge Distillation \cite{Hinton2015Distilling} outputs over-confident prediction for the correct category and low confidence for incorrect classes.
However, this category similarity information is exactly the most important knowledge that should be transferred to the student. For example, the probability of incorrectly classifying a cat as a car must be lower than the probability of misclassifying it as a dog. Thus, we argue that the image labels are more informative for the specific images. In other words, the teacher should guide the student to distinguish what the image is and what it looks like. 

In this work, we employ the mixup \cite{DBLP:conf/iclr/ZhangCDL18} to create our virtual training samples by a weighted linear interpolation. In this way, the virtual samples with mix-labels can provide additional intra- and inter-class relations in datasets. In contrast to existing approaches, the proposed method transfers the similarity correlation between different instances instead of similarities information of individual samples. It means that we force the virtual samples created by multiple mages to produce similar probability distribution in the teacher and student networks. Due to the soften probability distribution from virtual samples, the teacher's knowledge could be easily learned by the compact student model that further improve its robustness.

To sum up, our contributions in this paper can be summarized as follow:

\begin{itemize}
\item We propose similarity transfer for knowledge distillation, which utilizes the valuable knowledge of similarities between categories on multiple instances.

\item We introduce mixup technique to create virtual samples and a novel knowledge distillation loss to combine the mixed labels.

\item Extensive experiments are conducted on various public datasets such as CIFAR-10/100, CINIC-10 and Tiny-ImageNet. We experimentally demonstrate that our approach significantly outperforms the vanilla knowledge distillation and other SOTA approaches.
\end{itemize}

The rest of this paper is organized as follows. Related work is reviewed in Part II. And we present the proposed similarity transfer for knowledge distillation architecture in Part III. Experimental results are presented in Part IV. Finally, Part V concludes this paper.

\section{Related Works}
In this section, we first briefly introduce the backgrounds of network model compression and acceleration and then summarize existing works on knowledge distillation. Finally, we review existing works related to the mixup technique. 

\subsection{Model compression and Acceleration.}
Recently, CNNs become predominant in many computer vision tasks. Network architectures become deeper and wider which achieve better performances than the shallow ones. However, these complicated networks also bring in high computation and memory costs, which obstruct us to deploy these models into real-time applications. Thus, model compression and acceleration aim to reduce the model size and computation cost meanwhile maintaining high performance. One kind of method directly designs small network structure by modifications on convolution since the original deep model has many redundant parameters. For example, MobileNet \cite{DBLP:journals/corr/HowardZCKWWAA17} introduces depth-wise separable convolution to build block instead of standard convolution. ShuffleNet \cite{zhang2018shufflenet} proposes point-wise group convolution and channel shuffle to maintain a decent performance without adding the burden of computation. These approaches could retain a decent performance without adding too much computing burden at inference phase.

Besides, there has been another kind of method which attempts to remove the redundant information of teacher network. For example, network pruning aims to boost the speed of inference by getting rid of redundancy of the large CNN model. Han et al. \cite{DBLP:conf/nips/HanPTD15} propose to boost the speed of inference by deleting unimportant connections. However, network pruning methods require many iterations to converge and the pruning threshold needs to be set manually. Network decomposition use matrix decomposition technique to decompose the original convolution kernel in CNNs model. For example, Novikov et al. \cite{DBLP:conf/nips/NovikovPOV15} propose to reduce the number of parameters while preserve the original performance of network by converting the dense weight matrices of the fully-connected layers to the Tensor Train format. Network quantization methods aim to represent the ﬂoat weights with fewer bits. Yang et al. \cite{DBLP:conf/cvpr/YangSXTLDH019} propose to simulate the quantization process with a differentiable function. Note that, above mentioned works can be combined with KD methods for further improvement.

\begin{figure*}
\centerline{\includegraphics[width=1\textwidth]{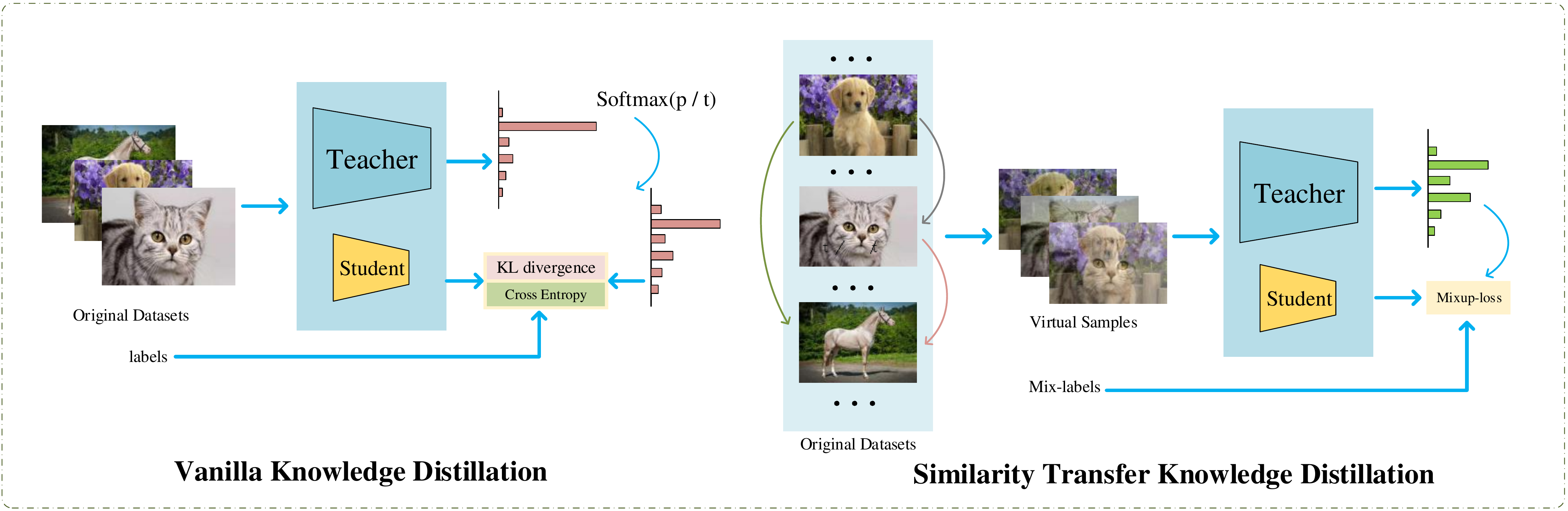}}
\caption{The whole framework of our Similarity Transfer Knowledge Distillation vs. Vanilla Knowledge Distillation. STKD extracts more similarities information among categories by feeding the virtual samples with original samples of mixed labels. As shown in the right, STKD outputs more soften probability distributions from the teacher and transfers the similarities information to the student.}\label{fig:framwork}
\end{figure*}

\subsection{Knowledge Distillation.} 
Different from above methods, knowledge distillation enrich and get the student model by extracting kinds of knowledge from the fixed teacher model. To address the challenge of deploying CNNs in resource-constrained edge devices, Bucilua et al. \cite{DBLP:conf/kdd/BucilaCN06} first propose to transfer the knowledge of an ensemble of models to a small model. Then Caruana et al. \cite{DBLP:conf/nips/BaC14} propose to train student model by mimicking the teacher model's logits. Later, Hinton et al.  \cite{Hinton2015Distilling} popularize the idea of knowledge distillation, which efficiently transfers knowledge from large teacher network to compact student network by mimicking the class probabilities outputs. Note that, the outputs of the teacher network are defined as the dark knowledge in KD, which provides similarity information as extra supervisions compared with one-hot labels. In other words, there are two sources of supervision used to train the student in KD, one from the ground-truth label, and the other from the soft targets of teacher network.

Afterward, some recent works \cite{Romero2015FitNets}~\cite{Zagoruyko2016Paying} extend KD by distilling knowledge from intermediate feature representations instead of soft labels.
For example, FitNets \cite{Romero2015FitNets} propose to train student network by mimicking the intermediate feature maps of teacher network, which are defined as hints. Inspired by this, Zagoruyko et al. ~\cite{Zagoruyko2016Paying} propose to match the attention maps between the teacher and the student, which are defined from the original feature maps as knowledge. Wang et al. \cite{DBLP:journals/corr/HuangW17a} propose to improve the performance of student network by matching the distributions of spatial neuron activations between the teacher and the student. Recently, Heo et al. \cite{DBLP:conf/aaai/HeoLY019a} introduce the activation boundary of the hidden neuron as knowledge for distilling the compact student network.

However, the aforementioned knowledge distillation methods only utilize the knowledge contained in the output of specific layers of the teacher network. More richer knowledge between different layers is explored and utilized for knowledge distillation. For example, Yim et al. \cite{Yim2017A} propose to use Gram matrix between different feature layers as distilled knowledge, which named flow of solution process (FSP) that reflects the relations of different features maps. Lee et al.  \cite{DBLP:conf/eccv/LeeKS18} propose to extract valuable correlation information as knowledge between different feature maps using singular value decomposition (SVD). However, these methods only focus on the knowledge contained in the individual data samples and ignore the similarities between categories of multiple samples, which also valuable for knowledge distillation.

Thus, some very recent works \cite{park2019relational}\cite{tung2019similarity} aim to explore the relationship between data samples. Park et al. \cite{park2019relational} propose a relation knowledge distillation (RKD) method, which extracts mutual relations of data samples by the proposed distance-wise and angle-wise distillation losses. Tung et al. propose a similarity preserving (SP) knowledge distillation method which transfers the similar activations of input pairs from the teacher network to the student network. Peng et al. \cite{DBLP:conf/iccv/PengJLZWLZ019} propose to capture the high order correlation between samples using the kernel method. However, the similarities information between instances is hard to learn for the student network due to the high dimension in the embedding space. Different from existing works, the proposed approach presents the similarity correlation between instances through the probability distribution outputs of virtual samples, which are created by the mixup technique. In contrast to the vanilla KD which softens the softmax outputs using a temperature hyperparameter, our approach directly outputs a softened probability distribution using the virtual samples.

\subsection{Mixed Samples Data Augmentation.} 
There are many training strategies to further improve the performance of CNNs, such as data augmentation and regularization techniques. In particular, mixed sample data augmentation has gain increasing attention due to its state-of-the-art performance. It trains the models using the augmented data set by combining data samples according to some policy.

For example, Zhang et al. \cite{DBLP:conf/iclr/ZhangCDL18} first introduce a data augmentation approach named mixup, which could improve the generalization of CNNs by linearly interpolating a random pair of original training data to create a new training data 
and corresponding labels. Inspired by this, some mixup variants \cite{DBLP:conf/icml/VermaLBNMLB19}\cite{DBLP:conf/wacv/SummersD19} \cite{DBLP:conf/acml/TakahashiMU18} have been proposed. Later, Cutmix \cite{DBLP:conf/iccv/YunHCOYC19} proposes to replace the removed regions with a patch from another image. FMix \cite{DBLP:journals/corr/abs-2002-12047} proposes to use binary masks obtained by applying a threshold to low frequency images sampled from Fourier space. Following them, Pairing Samples \cite{DBLP:journals/corr/abs-1801-02929} propose to take an average of two samples for each pixel. Recently, Cubuk et al. \cite{DBLP:journals/corr/abs-1805-09501} employ automatically searching to improve data augmentation policies that called AutoAugment. Although these methods have achieved remarkable performance to the training of CNNs, they suffer from the prohibitive training time.

\section{Proposed Method}
In this section, we first revisit the definition of vanilla KD \cite{Hinton2015Distilling}, which transfers knowledge from the high-capacity teacher network to student network with soft labels. Then we describe the proposed similarity transfer for knowledge distillation. Fig. \ref{fig:framwork} compares the vanilla KD with STKD and describes the whole framework of our approach.

\subsection{Problem Definition}
In our approach, we aim to explore a more effective knowledge, which contains in the similarity correlation between instances, to guide the training of student model.
Moreover, we employ mixup technique to represent these knowledge by introducing virtual samples. 

For simplicity, we define the large, complex \textsl{Teacher} deep neural network as \textsl{T} with learned parameters $P_{T}$, the less complex \textsl{Student} network as \textsl{S} with learned parameters $P_{S}$. The original training data consist of tuples of input data and target $(x,y) \in D$. In general, the conventional knowledge distillation methods train the \textsl{Student} by minimizing the following objective function with respect to the parameters $P_{S}$ over the training samples $(x,y) \in D$:

\begin{equation}
\label{eq:1}
\mathcal{L} = {\mathcal{L}_{KD}}(S(x,P_{S}),T(x,P_{T})) + \lambda\mathcal{H}(y_{s},y)
\end{equation}

\noindent where $\lambda$ is a balancing hyperparameter,
$\mathcal{H}$ is the cross-entropy loss that computes on the predicted label $y_{s}$ and the corresponding ground truth labels $y$. And $\mathcal{L}_{KD}$ is the distillation loss such as cross-entropy or mean square error. $T(x,P_{T})$ and $S(x,P_{S})$ represent the softmax output of the $Teacher$ and $Student$ respectively.

For example, Hinton et al. \cite{Hinton2015Distilling} use pre-softmax outputs for $T(x,P_{T})$ and $S(x,P_{S})$, and Kullback-Leibler divergence for $\mathcal{L}_{KD}$:

\begin{equation}
\sum_{x_{i}\in D} KL ( softmax(\frac{T(x_{i},P_{T})}{ t }) , softmax(\frac{S(x_{i},P_{S})}{ t }) )
\end{equation}

\noindent where $t$ is a temperature hyperparameter to soften the softmax outputs of teacher network. Thus, we 
need to set a proper $t$ manually. However, it is hard for us to obtain a proper temperature $t$ due to the gap between the teacher and student networks.

Likewise, other works \cite{Romero2015FitNets}\cite{Yim2017A} involved an objective function that can also be formulated as a form of Eq. \ref{eq:1}. However, these conventional KD methods only focus on the similarity correlation between categories of individual samples while ignore the similarity correlation between different samples. Moreover, the temperature hyperparameter $t$ need to be set manually which is used to soften the probability  distribution. 

\subsection{Similarity Transfer Knowledge Distillation}
In this section, we introduce similarity transfer for knowledge distillation in detail. Usually, traditional knowledge distillation methods \cite{Hinton2015Distilling} utilize the temperature hyperparameter to soften the one-hot labels to soft labels. Different from existing works, STKD presents the similarity correlation between different samples through the probability distribution outputs of virtual samples, which are created by the mixup technique. 

To be specific, we first employ the recently proposed mixup \cite{DBLP:conf/iclr/ZhangCDL18} to create the virtual samples. We randomly sample two samples $(x_{i},y_{i})$ and $(x_{j},y_{j})$ from $D$. Then it generates the new synthetic sample $(x,y)$ by a weighted linear interpolation of these two samples:

\begin{equation}
\label{eq:03}
x = \lambda x_{i} + (1-\lambda) x_{j}
\end{equation}
\begin{equation}
\label{eq:04}
y = \lambda y_{i} + (1-\lambda) y_{j}
\end{equation}

\noindent where the merging coefficient $\lambda \in [0,1]$ is a random number drawn from the $Beta(\alpha ,\alpha)$ distribution. And $y$ is the label of $x$ which is a convexed combination of the labels from $x_{i}$ and $x_{j}$.

As can be seen from Figure. \ref{fig:framwork}, the virtual samples are created by this mixup technique. In other words, each sample from the virtual set is formed by two arbitrary original images by a weighted linear interpolation.

By Eq. \ref{eq:04}, we turn the one-hot labels to mixed labels, which also contribute to the similarity information between different samples. Thus, the mixed labels of the virtual samples are employed in our framework. Therefore, the mix loss can be formulated as:

\begin{equation}
\mathcal{L}_{mix} = \lambda\mathcal{H}(y_{s_{i}},y) + (1-\lambda) \mathcal{H}(y_{s_{j}},y)
\end{equation}

\noindent where $y_{s_{i}}$ and $y_{s_{j}}$ are the predicted labels corresponding the ground truth labels $y$. And $\lambda$ is a hyperparameter as same as the coefficient in the mixup technique. Note that, we supervise the training of student network by using these mixed labels and the teacher network's logits.

Then we distill the similarity knowledge between multiple samples from teacher network to student network using the created virtual samples. Here we define the total loss function for similarity transfer knowledge distillation as follows:

\begin{eqnarray}
\label{eq:2}
\mathcal{L}_{total} = \lambda\mathcal{H}(y_{s_{i}},y) + (1-\lambda) \mathcal{H}(y_{s_{j}},y) +\\\nonumber 
{\mathcal{L}_{KD}}(S(x,P_{S}),T(x,P_{T}))
\end{eqnarray}

Different from the vanilla KD method, we get rid of the temperature hyperparameter in our method. Thus, the probability distribution output of $T$ and $S$ is defined as $T(x,P_{T}) =  softmax(T(x_{i},P_{T})$ and $S(x,P_{S}) =  softmax(S(x_{i},P_{S})$ respectively. Note that the classification loss has the hyper-parameter $\lambda$ to balance the weight of the mix labels.

\begin{algorithm}[t]
\caption{Similarity Transfer Knowledge Distillation} 
\label{con:algorithm}
\hspace*{-0.08cm} {\bf Input:} 
A pre-trained teacher model $T$ with parameters $P_{T}$.\\
\hspace*{-0.08cm} {\bf Input:} 
The original training data $(x,y) \in D$.\\
\hspace*{-0.08cm} {\bf Input:} 
Hyper-parameter (learning rate, coefficient $\lambda$, etc).\\
\hspace*{-0.08cm} {\bf Output:} 
A compact student model $S$ with parameters $P_{S}$.

\begin{algorithmic}[1]
\State {Initialize: The student network $S$ and training hyper-parameters.}
\State \bfseries Repeat: \mdseries
    \State \quad \bfseries Stage 1: Creating the Virtual Samples by Mixup. \mdseries
    \State \quad Sample a batch $(x,y)$ from the training dataset.
    \State \quad Sample two samples $(x_{i},y_{i})$ and $(x_{j},y_{j})$ randomly 
    \State \quad from the batch.
    \State \quad Synthetic the virtual samples by Eq. \ref{eq:03} and \ref{eq:04}.
    
    \State \quad \bfseries Stage 2: Training the student model. \mdseries
    \State \quad Sample a batch $(x,y_{i},y_{j})$ from the virtual samples.
    \State \quad Calculate the softmax outputs of teacher network and 
    \State \quad student network: $P_{t}(x)$, $P_{s}(x)$ 
    \State \quad Calculate the knowledge distillation loss in Eq. \ref{eq:2}.
    \State \quad Update weights in $P_{S}$ according to the gradient.
    
\State \bfseries Until: \mdseries convergence.
\end{algorithmic}
\end{algorithm}

Unlike previous methods, we define the similarity correlation between instances as knowledge and present it through probability distribution of the virtual images. To this end, we employ the mixup technique to create virtual samples which is effective for our method. It generates mixup images that contain multiple images' similarity correlation on the one hand, on the other hand also 
regularizes the student model through knowledge distillation to favor simple linear behavior in-between training samples. Moreover, we could generate different mixup images by varying the coefficient $\lambda$. We study this coefficient in the ablation study.

In this way, teacher network transfers more valuable information to the student network than existing methods. That means the teacher's dark knowledge could be better transferred to the student.

\subsection{Training Procedure}

Algorithm \ref{con:algorithm} describes the details of the whole training paradigm. We first obtain a pre-trained teacher model $T$ with parameters $P_{T}$  which is trained using standard back-propagation on original dataset $(x,y) \in D$. 

Then we iterate the following stages until the accuracy of the student $S$ converges. 1) We create the virtual set instead of original dataset using the mixup technique. Specifically, we use a single data loader to obtain the mini-batch and apply the mixup to the same mini-batch after random shuffling. 2) We feed these virtual samples to our knowledge distillation framework, which distills the similarity correlation between instances from the pre-trained teacher to the student network using our STKD loss in Eq. \ref{eq:2}.

Fig. \ref{fig:framwork} illustrates the overall framework of STKD.  Different from the previous methods, the proposed STKD regularizes the student network to favor the simple linear behavior under the teacher's guidance. In other words, we employ the mixup technique to provide more similarity information among classes and make the high dimensional representation of teacher more easily transferable to the student. We will demonstrate its effectiveness in the following section.

\section{Experiments and Results}
\label{sec:guidelines}

In this section, extensive experiments are conducted to verify the effectiveness of the proposed approach. We first design comparison experiments with state-of-the-art (SOTA) KD methods (e.g., AT \cite{Zagoruyko2016Paying}, SP \cite{tung2019similarity} and RKD \cite{park2019relational}) to test our approach. Additionally, different network architectures (e.g., ResNet \cite{he2016deep}, WideResNet \cite{Zagoruyko2016Wide}, VGG \cite{simonyan2014very}) are explored in ablation studies. 

Note that, we implement the experiments using PyTorch on Nvidia 1080TI GPU devices. For all the experiments, the results are averaged over 5 trials of different randomly selected seeds.

\subsection{Experiment Settings}
\noindent\textbf{Datasets.} To demonstrate our method under general situations of data diversity, we run experiments on CIFAR-10/100, CINIC-10 and Tiny-ImageNet which are popular benchmark datasets for image classification. CIFAR-10 \cite{krizhevsky2009learning} and CIFAR-100 both contain 50K training images and 10K test images. CIFAR-10 has 10 categories when CIFAR-100 contains 100 classes. The CINIC-10 dataset consists of images from both CIFAR-10 dataset and ImageNet dataset, whose scale is closer to ImageNet. It is composed of 270,000 images at a spatial resolution of $32\times32$ via the addition of down-sampled ImageNet images. We adopt the CINIC-10 dataset for rapid experimentation. Tiny-ImageNet dataset \cite{le2015tiny} is a popular subset of the ImageNet dataset \cite{Russakovsky2015ImageNet}. It contains 100k training images with $64\times64$ resolution in 200 categories, 10K additional images for testing. Each class has 500 training images, 50 validation images, and 50 test images.

\noindent\textbf{Network architecture.} We use three state-of-the-art convolutional neural network architectures: ResNet \cite{DBLP:conf/cvpr/HeZRS16}, Wide Residual Network (WRN) \cite{DBLP:conf/bmvc/ZagoruykoK16} and VGG \cite{DBLP:journals/corr/SimonyanZ14a}. For each network family, we employ a deep or wide one as the teacher network and a shallow or thin one as the student network. Note that, WRN has a standard convolutional layer followed by three groups of residual blocks, each of size n. Additionally, it uses an additional widen factor $m$ to increase the width, which could bring more representation ability. We denote WRN as WRN-n-m in our experiment. In the following experiments, we take ResNet-110, WRN-40-2 and VGG-13 as the teacher networks, ResNet-20, WRN-40-1 and VGG-8 as student networks. In addition, we use different network architectures (e.g., ResNet-50 as teacher and VGG-8 as student) in ablation studies.

\noindent\textbf{Evaluation Metric.} We adopt the classification accuracy as the evaluation metric. Note that, the accuracy is computed as median of 5 runs with different seeds.

\begin{table*}
\begin{center} 
\caption{Classification accuracy (\%) on CIFAR-10 and CIFAR-100 datasets (5 runs). Baseline means the student network trains individually. STKD means the classification accuracy in our method.}
\label{table:1}
\begin{tabular}{c|c|c|c|c|c|c|c||c}
\hline
Dataset & Model(S/T) & Baseline &  KD \cite{Hinton2015Distilling}  & AT \cite{Zagoruyko2016Paying}& SP \cite{tung2019similarity}& RKD \cite{park2019relational}& STKD & Teacher\\
\hline
\hline
\multirow{3}{*}{CIFAR-10} 
&\tabincell{c}{WRN-40-1 (0.56M)\\ WRN-40-2 (2.20M)} & 93.84 & 94.17 & 93.96&94.21 &94.06&\bfseries94.44\mdseries &94.94  \\
\cline{2-9}
&\tabincell{c}{ResNet-20 (0.27M)\\ ResNet-110 (1.7M)} & 92.56 & 93.08 & 93.21 & 93.30 & 93.20&\bfseries93.52\mdseries &94.51 \\
\cline{2-9}
&\tabincell{c}{VGG-8 (0.27M)\\ VGG-13 (1.7M)} & 92.76 & 93.14 & 93.29 &93.08 &93.31 &\bfseries93.68\mdseries &94.27 \\
\cline{2-9}

\hline
\hline
\multirow{3}{*}{CIFAR-100} 
&\tabincell{c}{WRN-40-1 (0.56M)\\ WRN-40-2 (2.20M)} & 72.08 & 73.34 & 73.77 & 73.46& 73.08&\bfseries74.63\mdseries & 75.61 \\
\cline{2-9}
&\tabincell{c}{ResNet-20 (0.27M)\\ ResNet-110 (1.7M)} & 69.14 & 70.69 & 70.97 & 71.02& 70.77&\bfseries72.14\mdseries & 74.37\\
\cline{2-9}
&\tabincell{c}{VGG-8 (0.27M)\\ VGG-13 (1.7M)} & 70.39 & 72.87 & 73.43 & 73.49 & 73.01&\bfseries 73.65\mdseries & 74.66\\
\cline{2-9}
\hline
\end{tabular}
\end{center}
\end{table*}

\begin{figure}
\centerline{\includegraphics[width=\columnwidth]{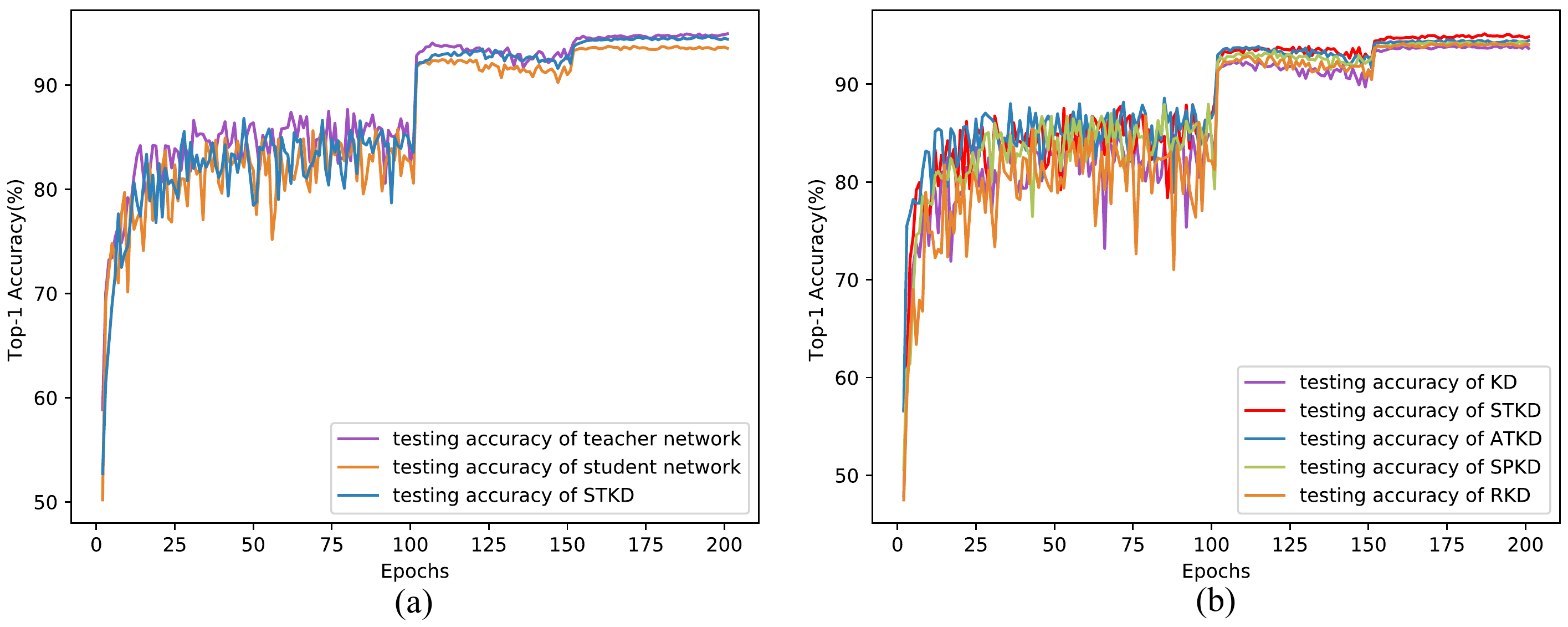}}
\caption{(a) Testing accuracy of the pre-trained teacher, student from our method and student trains individually. (b) Testing accuracy of different knowledge distillation methods on CIFAR-10.}\label{fig:loss}
\end{figure}

\subsection{Experiment on CIFAR-10/CIFAR-100}
 
On CIFAR dataset, we use three groups of teacher/student model pairs including ResNet-110/ResNet-20, WRN-40-2/WRN-40-1 and VGG-13/VGG-8. We apply a standard horizontal ﬂip and random crop data augmentation scheme which is widely adopted for these datasets. We set the weight decay to $10^{-4}$., batch size to 128, and use stochastic gradient descent (SGD) with momentum 0.9. The initial learning rate is set to $\gamma$ = 0.1 and divided by 10 at 80th, 100th, 150th epochs, totally 200 epochs. We use the same setting for training the teacher network and distilling the student network.

We conduct baseline comparisons with respect to the vanilla KD \cite{Hinton2015Distilling}, AT \cite{Zagoruyko2016Paying}, SP \cite{tung2019similarity} and RKD \cite{park2019relational}. For vanilla KD, we set temperature hyperparameter $t = 4$ following the experiments in \cite{Zagoruyko2016Paying}. We set the hyperparameter $\beta = 1000$ for AT, $\gamma = 0.1$ for SP respectively.

Table \ref{table:1} summarizes the classification accuracy of several teacher/student model pairs on CIFAR-10/100 datasets. For WRN network architecture, the teacher and student networks have the same depth but different width (WRN-40-2 teacher with WRN-40-1 student). 

Our method gets 94.44\%, 93.52\% and 93.68\% of top-1 accuracy for WRN-40-1, ResNet-20 and VGG-8 on CIFAR-10, respectively. It substantially surpasses the vanilla KD by 1.6\%, 1,9\% and 1.8\%. Compared to other baselines, the student network from STKD still significantly surpass the three related SOTA methods. For CIFAR-100, our method also has 1.6\%, 1.7\% and 1.8\% top-1 accuracy promoting over vanilla KD. Moreover, our approach still surpasses the three SOTA distillation related methods, which verifies the effectiveness of our approach. These results validate our intuition that the similarity correlation between instances encodes valuable knowledge from the teacher network and provides an important supervision for knowledge distillation.

Figure \ref{fig:loss} plots the accuracy change curves for WRN-40-2/WRN-40-1 experiments on CIFAR-10. As can be seen from Figure \ref{fig:loss} (a), it shows the testing accuracy curves of teacher network, student network which is trained individually and student network from our method. Note that, STKD gets a significant improvement compared to the student trained individually and its final accuracy value is close to that of the teacher. Figure \ref{fig:loss} (b) shows the validation accuracy change curves of WRN-40-1 over time among different knowledge distillation approaches on CIFAR-10 dataset. We can observe that our method performs a significant improvement on the final accuracy and outperforms the other SOTA approaches.

From the perspective of the teacher model, our method also shows the potential to compress large networks into more compact ones with minimal accuracy loss. For example, we distill the knowledge from the pre-trained WRN-40-2 teacher network, which contains $2.20M$ parameters,  to a much smaller WRN-40-1 student network, which contains $0.56M$ parameters. The student network gets a 4$\times$ compression rate with only $0.5\%$ loss in classification accuracy using off-the-shelf Pytorch.

\begin{table}
\begin{center}
\caption{Classification accuracy (\%) on CINIC-10 (5 runs). WRN-40-2 is as the teacher network, WRN-40-1 as the student network. Baseline means the WRN-40-1 trains individually. STKD means the WRN-40-1 results in our method. }
\label{table:2}
\begin{tabular}{c|c|c|c}
\hline
Type & Model & Params(M) & Acc (\%)\\
\hline
\hline
Baseline & WRN-40-1 & 0.56 & 84.30  \\
KD &  WRN-40-1 & 0.56  &  85.04\\
AT &  WRN-40-1 & 0.56 &  85.53\\
RKD &  WRN-40-1 & 0.56 &  84.96\\
SP &  WRN-40-1 & 0.56 &  85.16\\
STKD &  WRN-40-1 & 0.56 & \bfseries85.94\mdseries \\
\hline
\hline
Teacher &  WRN-40-2 & 2.20 & 86.40 \\
\hline
\end{tabular}
\end{center}
\end{table}

\subsection{Experiment on CINIC-10}

In this section, we conduct experiments on CINIC-10 dataset, whose scale is closer to that of ImageNet. We use WRN-40-2 and WRN-40-1 as the teacher network and student network respectively. For training phase, we apply CIFAR-style data augmentation with horizontal ﬂips and random crops. The beginning learning rate is set to 0.01 and decayed by a factor of 10 after the 100th, 160th and 220th epochs. All the networks are trained for 240 epochs using SGD with Nesterov momentum, a batch size of 96. 

For fair comparison, we adopt the same setting as mentioned above for all baseline methods. We set the hyperparameters for the baseline methods ($t = 8$ for KD; $\beta = 100$ for AT, $\gamma = 0.1$ for SP). The results on CINIC-10 are shown in Table \ref{table:2}. Our method achieves a 85.94\% classification accuracy, which surpasses the student network trained individually by promoting 1.64\%. Moreover, STKD substantially surpasses the vanilla KD by 0.9\%. It also shows comparable performance with other SOTA approaches. These results imply that STKD method induces more meaningful dark knowledge such as the similarity information between instances than other baseline methods.

\begin{table}
\begin{center}
\caption{Classification accuracy (\%) on Tiny ImageNet (5 runs). Baseline means the WRN-40-1 trains individually. STKD means the WRN-40-1 results in our method.}
\label{table:3}
\begin{tabular}{c|c|c|c}
\hline
Type & Model & Params(M) & Acc (\%)\\
\hline
\hline
Baseline & WRN-40-1 & 0.56 &  56.93\\
KD &  WRN-40-1 & 0.56  &  58.52\\
AT &  WRN-40-1 & 0.56 &  58.43\\
RKD &  WRN-40-1 & 0.56 &  57.02\\
STKD &  WRN-40-1 & 0.56 & \bfseries 59.25\mdseries \\
\hline
\hline
Teacher &  WRN-40-2 & 2.20 &  62.21\\
\hline
\end{tabular}
\end{center}
\end{table}

\subsection{Experiment on Tiny-ImageNet}
For rapid experimentation, we conduct experiments on Tiny-ImageNet dataset, which is a popular subset of the ImageNet. We adopt the VGG network architecture for the following experiments. To be specific, VGG-13 is employed for the teacher network, which achieves 62.21\% classification accuracy. VGG-8 is used for the student network. Both of the networks are trained for 240 epochs.

In our Tiny ImageNet classification experiments, we apply random rotation and horizontal flipping for data augmentation. We optimize the model using stochastic gradient descent(SGD) with mini-batch 128 and momentum 0.9. The learning rate starts from 0.1 and is multiplied by 0.2 at 60, 120, 160, 200, 250 epochs. We totally train the network for 300 epochs and adopt the deep and wide WRN (WRN-40-1) for a teacher model and WRN-16-1 as a student model.

In this section, ResNet with four blocks is selected with channel size of 16, 32, 64 and 128. Random crop and horizontal flip are used for data augmentation. 
Table \ref{table:3} shows the results, where baseline denotes the student network trained individually. As can be seen from Table \ref{table:3}, our method, as well as the other SOTA methods, gets higher classification accuracy than the original student network. Moreover, our STKD method achieves better performance than the KD and shows comparable performance with the other knowledge distillation methods, and further overcomes them.

\subsection{Ablation Study}

\textbf{Evaluation on Different Network Architectures.} To further evaluate the performance of STKD, we perform additional experiments on different network architecture families for the student/teacher pairs on CIFAR-100. Table \ref{table:4} shows the experimental results. 

As can be seen from Table \ref{table:4}, STKD continuously outperforms KD and other SOAT methods, which indicates the effectiveness of our method. In particular, STKD achieves the accuracy of 69.07\%, 73.81\% and 68.22\% for three network pairs of different network families respectively, obtaining a performance gain of 1.07\%, 0.92\% and 2.33\% over SP. However, we observe that some methods such as AT \cite{Zagoruyko2016Paying} perform quite poorly when the teacher and student belong to different network architectures. Thus, the results validate that our approach is robust to teacher/student pairs. And we conclude that the knowledge contained in the similarity correlation between different instances helps to improve the robustness of student network.

\begin{table}
\begin{center}
\caption{Additional experiments under different network architecture families on CIFAR-100. Classification accuracy (\%) over five runs is reported. The best performance is shown in bold. Baseline means the student network trained individually. STKD means the performance of student network in our method.}
\label{table:4}
\begin{tabular}{c|c|c|c}
\hline
\tabincell{c}{Teacher\\ Student} & \tabincell{c}{ResNet-50\\ MobileNetV2} & \tabincell{c}{ResNet-50\\ VGG-8} & \tabincell{c}{VGG-13\\ MobileNetV2}\\
\hline
\hline
Baseline & 64.64 & 70.40 & 64.64  \\
KD & 67.62  & 73.39  &  67.36\\
AT & 58.58  & 71.14 &  59.83\\
SP & 68.00  & 72.89 &  65.89\\

STKD & \bfseries69.07\mdseries  & \bfseries73.81\mdseries & \bfseries68.22\mdseries \\
\hline
\hline
Teacher & 79.34  & 79.34 & 74.31 \\
\hline
\end{tabular}
\end{center}
\end{table}

\textbf{Impact of Different coefficients for mixup.} In this subsection, we explore the impact of coefficients for mixup. We adopt ResNet-110 as the teacher network, ResNet-20 as the student network. The results on CIFAR-100 could be found in Table \ref{table:5}, which illustrates how the performance of STKD is affected by the choice of the coefficient $\lambda$. By setting different $\lambda$ for STKD, we observe that STKD achieves the accuracy of 72.14\% ($\lambda = 0.50$), which is the best performance in our settings. 

\begin{figure}
\centerline{\includegraphics[width=\columnwidth]{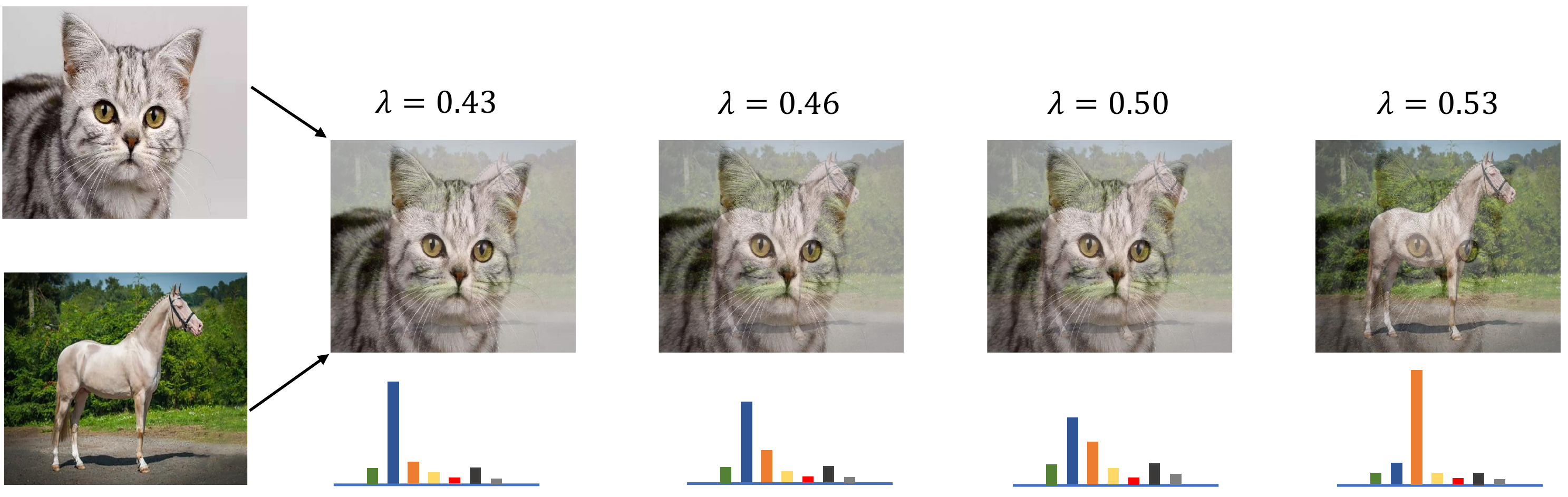}}
\caption{Different mixup images from the same sample pairs by varying the mixup coefficient $\lambda$. We plot the probability distribution of each mixup image from the teacher network.}\label{fig:mixupcoeffi}
\end{figure}

Furthermore, we show the mixup images from the same sample pairs by varying the mixup coefficient $\lambda$ in Figure \ref{fig:mixupcoeffi}. The outputs of probability distribution are also plotted under each mixup image. As can be seen from Figure \ref{fig:mixupcoeffi}, when $\lambda = 0.50$, the mixup image obtains a smoother probability distribution than other settings. It means that the relative probability assigned to similar categories encodes semantic similarity between similar categories. Note that, previous conventional KD methods usually achieve this goal by setting a large temperature hyperparameter $t$.

\begin{table}
\begin{center}
\caption{Classification accuracy (\%) on CIFAR-100 (5 runs). Baseline means the ResNet-20 trains individually. STKD means the ResNet-110 results in our method.}
\label{table:5}
\begin{tabular}{c|c|c|c|c}
\hline
$\lambda$ & 0.43 & 0.46 & 0.50 & 0.53\\

$1-\lambda$ & 0.57 & 0.54 & 0.50 & 0.47 \\
\hline
\hline
Baseline & 69.14 & 69.14 & 69.14 & 69.14 \\

STKD & 70.21  & 71.08 & \bfseries 72.14\mdseries & 69.82 \\
\hline
\hline
Teacher & 74.37  & 74.37 & 74.37 & 74.37 \\
\hline
\end{tabular}
\end{center}
\end{table}

\begin{figure}
\centerline{\includegraphics[width=\columnwidth]{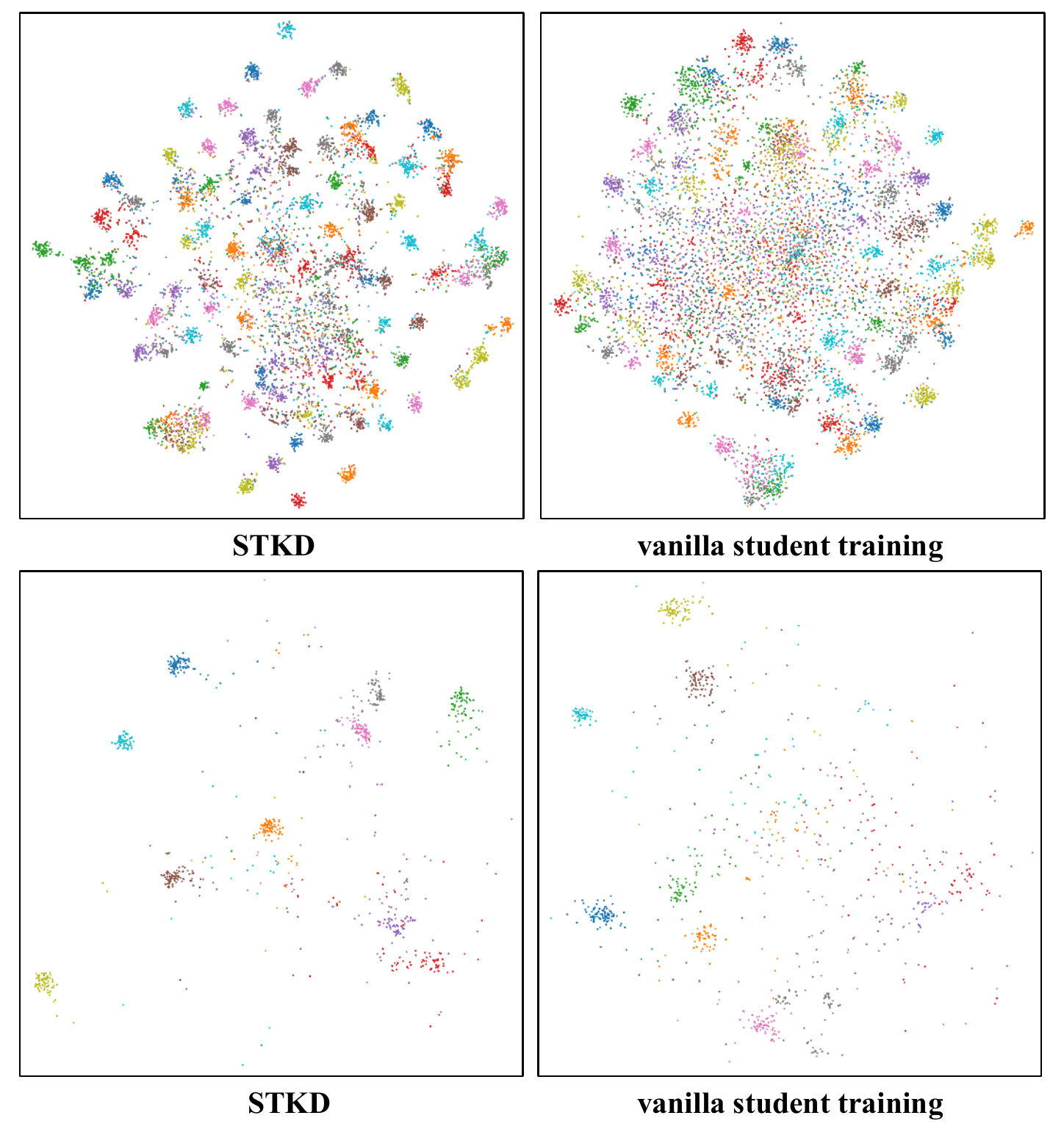}}
\caption{The tSNE visualization over 100 classes of the STKD (left) and the vanilla student training (right). Each color represents a category, best viewed in color. The first row shows 100 classed together. The second row shows 10 sampled classes from CIFAR-100.}\label{fig:visualize}
\end{figure}

\textbf{Visualization the distribution of student networks.} In this subsection, we visualize the output distribution of student from STKD and vanilla training methods on CIFAR-100. Figure \ref{fig:visualize} shows the visualization results over 100 classes and 10 sampled classes using tSNE \cite{DBLP:conf/mlsp/MaatenW12} respectively. Notably, the feature space of STKD is significantly more separable than it in vanilla training manner.

\section{Conclusion}

In this paper, we introduce a novel knowledge distillation method, called similarity transfer knowledge distillation that aims to utilize the similarity correlation between different instances. Moreover, we adopt the mixup technique to encode semantic similarity between categories, which assigns the relative probability to different categories from instances.
And the one-hot labels are replaced by the mix labels in virtual samples, created by the mixup technique. Our experiments demonstrate that the dark knowledge from the teacher model could be better transferred to the student model and improve the training outcomes of the student. We also show that our approach can be regarded as the regularized term for the student in knowledge distillation. The experiment evaluation on three benchmarks CIFAR-10/100, CINIC-10 and Tiny-ImageNet verify the effectiveness of our approach, which achieves the state-of-the-art accuracy for a variety of network architectures.

\bibliographystyle{IEEEtran}

\bibliography{IEEEexample}

\end{document}